\documentclass[letterpaper]{article}
\pdfoutput=1
\usepackage{aaai}
\usepackage{times}
\usepackage{relsize}
\usepackage{amsmath}
\usepackage{dl-macros}
\newcommand{\URL}[1]{\mbox{\relscale{0.9}\mdseries\sffamily\slshape #1}}
\newcommand{\LURL}[1]{\ensuremath{\mbox{\relscale{0.9}\mdseries\sffamily\slshape ex:#1}}}
\newcommand{\url}[1]{{\relscale{0.9}\mdseries\sffamily\slshape #1}}
\newcommand{\literal}[2]{{\tt "}#1{\tt "}\^{}\^{}\url{#2}}
\newtheorem{theorem}{Theorem}
\newtheorem{definition}{Definition}
\newcommand{\proof}{\\ \rm {\bf Proof: }}
\newcommand{\proofsketch}{\\ \rm {\bf Proof Sketch: }}
\newcounter{examplecounter}
\newcommand{\EXAMPLE}{\refstepcounter{examplecounter}(\arabic{examplecounter})}

\title{Using Description Logics for \\ RDF Constraint Checking and Closed-World Recognition}
\author{
Peter F. Patel-Schneider \\
Nuance Communications \\
1198 East Arques Avenue, Sunnyvale, California, U. S A. \\
pfpschneider@gmail.com
}

\copyrighttext{Extended version of a paper of the same name that
  will appear in AAAI-2015.  Copyright \copyright{} 2014,  Nuance Communications.}
\begin{document}

\maketitle
\begin{abstract}
RDF and Description Logics work in an open-world setting where absence of
information is not information about absence.  Nevertheless, Description
Logic axioms can be interpreted in a closed-world setting and in this
setting they can be used for both constraint checking and closed-world
recognition against information sources.  When the information sources are
expressed in well-behaved RDF or RDFS (i.e., RDF graphs interpreted in the
RDF or RDFS semantics) this constraint checking and closed-world recognition
is simple to describe.  Further this constraint checking can be implemented
as SPARQL querying and thus effectively performed.
\end{abstract}

There has recently been considerable attention paid to the problem of
validating RDF \cite{cyganiak-et-al:rdf-11-concepts} or RDFS information.  There are several
commercial systems that provide facilities for RDF validation, including
TopQuadrant's SPIN \cite{spin}
and Clark\&Parsia's Stardog ICV \cite{stardog-icv-spec,stardog-icv}.  
There are several proposals for specifying the desired form of RDF information,
such as Resource Shapes \cite{rdf-shapes}.
In 2013 W3C
held an RDF Validation Workshop \cite{rdf-validation-workshop} to gauge
interest in the area, and W3C has started a new working group on RDF
validation \cite{rdf-shapes-wg}. 

Just what, however, is RDF validation?  

Some accounts and systems (such as Stardog ICV) identify RDF validation with
satisfying integrity constraints, similar to checking database integrity
constraints.  In this account, there are conditions 
(constraints) 
placed on
instances of classes, such as requiring that every person has a name and an
address, both strings. The defining characteristic here is that explicit
information is needed to satisfy the integrity constraint.  To pass the
constraint that a person has a name it is necessary to provide a particular
string for the name of the person, and not just state that the person has some
unknown name.

Other accounts and proposals, such as 
OSLC Resource Shapes \cite{ryman-et-al:oslc-resource-shape} (used by the
Open Services for Lifecycle Collaboration community)
and ShEx \cite{shex-primer,shex-definition},
identify RDF validation more with recognition,
similar to determining whether an individual belongs to a Description Logic
\cite{dlhbe2r} description.  
For example one might define shape that requires a name and an
address and then ask which individuals satisfy the shape's constraint.
Here, in contrast to the previous situation, the
validation is divorced from any type information in the data.  Again, however, there is
the requirement that explicit information is needed to match the shape---a
particular name must be provided, not just information that there must be one.

As shown in this paper,
Description Logics can be used to provide the necessary framework for both
checking constraints and providing closed-world recognition facilities,
and thus cover most of what SPIN and ShEx provide.

Why then are there claims
\cite{ryman-et-al:oslc-resource-shape,fokoue-ryman:oslc-resource-shape}
that OWL \cite{motik-et-al:owl-ss-fs}---the Semantic Web Description Logic---is
inadequate for these purposes?  
There are several aspects of the standard view of
Description Logics that might not be consonant with constraints and the kind
of recognition that might be desired.  However, 
Description Logic syntax and semantics, and their instantiation in OWL,
can serve as the basis for RDF constraint checking and closed-world
recognition.  The only change required is to consider a closed-world
variation of the Description Logic semantics.  Then the development of RDF
constraint checking and closed-world recognition is easy.

\section{The Basic Idea}

\subsection{Closed-World Recognition}

Let's first look at recognition.  In recognition we want to determine whether a
particular node in an RDF graph matches some criteria.  For example, John
in the RDF graph\footnote{Turtle \cite{turtle-1.1}
  will be used for writing RDF graphs throughout this paper.  Prefix and
  base statements will generally be omitted.} 

\begin{quote}
\LURL{John} \URL{foaf:name} \literal{John}{xsd:string} . \hfill \EXAMPLE \label{G1} \\
\LURL{John} \URL{foaf:phone} \literal{+19085551212}{xsd:string} . \\
\LURL{John} \URL{exo:friend} \LURL{Bill\/} . \\
\LURL{John} \URL{exo:friend} \LURL{Willy} .
\end{quote}

\noindent matches the ShEx shape

\begin{quote}
  \{ \URL{foaf:name} \URL{xsd:string}, \hfill \EXAMPLE \label{S1} \\
  \hspace*{1ex} \URL{foaf:phone} \URL{xsd:string}, \\
  \hspace*{1ex} \URL{exo:friend} [2] \}
\end{quote}

\noindent because John has a string value for his name, a string value for
his phone, and two friends.

Determining whether an
individual belongs to a Description Logic concept is also recognition.
The Description Logic description\footnote{The abstract syntax
  \cite{dlhbe2r} for
  Description Logics---a compact but non-ASCII syntax---will be used
  throughout this paper.}
corresponding to the ShEx Shape \eqref{S1} is

\begin{quote}$
 \exactly{1}{\URL{foaf:name}} \;\AND\;
    \all{\URL{foaf:name}}{\URL{xsd:string}} \;\AND \hfill \EXAMPLE \label{D1} \\
 \exactly{1}{\URL{foaf:phone}} \;\AND\;
    \all{\URL{foaf:phone}}{\URL{xsd:string}} \;\AND \\
 \exactly{2}{\URL{exo:friend}}
$\end{quote}

\noindent So it seems that Description Logics can easily handle ShEx
recognition.   Although the ShEx syntax is somewhat more compact here, as ShEx
has constructs that combine number restrictions and value restrictions, the
Description Logic syntax is not verbose and is quite reasonable.

\medskip

However, John does not match \eqref{D1} in the standard semantics of Description
Logic.  This is precisely because in this standard reading, {\em and in
  RDF}, the absence of information is not information about absence.  In the
standard Description Logic reading, and also in RDF, John could have more
than one name as far as the information in the above RDF graph is concerned.
Many Description Logics (and, again, RDF too) also do not assume that
different names refer to different individuals.  So Bill and Willy could be
the same person as far as the information in the above RDF graph is
concerned.

It turns out that expressive Description Logics have facilities to
explicitly state information about absence and information about differences
and thus can be used to state complete information, at least on a local
level.  For example, if we add information to \eqref{G1} stating that John
has only one name and phone, that John's only friends are Bill and Willy,
{\em and} that Bill is not the same as Willy, as in

\begin{quote}$
\LURL{John} \;\in\; \atmost{1}{\URL{foaf:name}} \\
\LURL{John} \;\in\; \atmost{1}{\URL{foaf:phone}} \\
\LURL{John} \;\in\; \all{\URL{exo:friend}}{\nominal{\LURL{Bill\/},\LURL{Willy}}} \\
\notsame{\LURL{Bill\/}}{\LURL{Willy}}
$\end{quote}

\noindent then John does match \eqref{D1}.

So it is not that Description Logics (including OWL) do not perform
recognition as in ShEx, it is just that Description Logics do not make the
assumption that absence of information is information about absence.  
In expressive Description Logics (again including OWL) it is possible to
explicitly state what comes implicitly from the assumption that
absence of information is information about absence.

\medskip

However, suppose that we want to make this assumption generally?  We could
manually add a lot of axioms like the ones above, but this is both tedious
and error-prone, and thus not at all a viable solution.  Instead we can
proceed by making the assumption that if the truth of some fact cannot be
determined from the information given, then that fact is false.  This is
often called the closed world assumption or negation by failure, as the
failure to prove some fact is used to support its falsity.  There is a very
large body of work on this topic 
(see the Related Work section for pointers into 
this work)
and there are many tricky questions that
arise with respect to closure in any sophisticated formalism, and expressive
Description Logics (including OWL) are indeed sophisticated.
As well, reasoning in expressive Description Logics that also have closed
world facilities is extremely difficult, even in simple cases.

Fortunately RDF and RDFS are unsophisticated and inexpressive, so neither the tricky
questions nor the reasoning difficulties arise if all information comes in the form of RDF 
triples interpreted under the RDF or RDFS semantics
\cite{hayes-patel-schneider:rdf-11-semantics}.  The basic idea is to treat 
the triples (and their RDF or RDFS 
consequences, if desired) as completely describing the world.  In this treatment
\begin{enumerate}
\item 
if a triple is not present then it is false and
\item 
different IRIs denote different individuals.  
\end{enumerate}
This is precisely the same idea that underlies model
checking, where a model is a finite set of ground first-order facts and
everything else is false.
First-order inference is undecidable, but determining whether a 
first-order sentence is true in one particular model (model checking) is much,
much easier.

In this way it is possible to use the Description Logic syntactic and semantic
machinery to define how to recognize descriptions under the same assumptions
that underlie ShEx.  The only change from the standard Description Logic
setup is to define how to go from an RDF graph to the Description Logic
model that the RDF graph embodies.  Definitions, even recursive definitions,
can also be handled.

This is all quite easy and conforms to a common thread of both theoretical
and practical work.  It also matches the theoretical underpinning of Stardog
ICV \cite{stardog-icv-spec,stardog-icv}.  Further, the approach can be implemented by
translation into SPARQL queries, showing that it is practical.  (There may
be some constructs of very expressive description logics that do not
translate into SPARQL queries when working with complete information, but at
least the Description Logic constructs that correspond to the usual
recognition conditions do so translate.)

\subsection{Constraint Checking}

Constraint checking does not appear to be part of the services provided by
Description Logics.  This has lead to claims that OWL cannot be used for constraint
checking.  However inference, which is the core service provided by
Description Logics, and constraint checking are indeed very closely related.

\begin{figure*}
\caption{Data for Example}\label{data}
\centering
\begin{tabular}{|l|l|l|}
\hline
\LURL{Amy} \URL{rdf:type} \URL{exo:UniStudent} . &
\LURL{Amy} \URL{exo:enrolled} \LURL{SUNYOrange} . &
\LURL{Amy} \URL{exo:friend} \LURL{Bill} . 
\\
\LURL{Amy} \URL{foaf:name} \literal{Amy}{xsd:string} . &
\LURL{Bill} \URL{exo:enrolled} \LURL{ReindeerPoly} . &
\LURL{Amy} \URL{exo:friend} \LURL{John} . 
\\
\LURL{Bill} \URL{rdf:type} \URL{exo:UniStudent} . &
\LURL{Bill} \URL{exo:enrolled} \LURL{HudsonValley} . &
\LURL{Bill} \URL{exo:friend} \LURL{Amy} . 
\\
\LURL{Bill} \URL{foaf:name} \literal{Bill}{xsd:string} . &
\LURL{John} \URL{exo:enrolled} \LURL{ReindeerPoly} . &
\LURL{Bill} \URL{exo:friend} \LURL{John} . 
\\
\LURL{John} \URL{rdf:type} \URL{exo:GrStudent} . &
\LURL{Susan} \URL{exo:enrolled} \LURL{ReindeerPoly} . &
\LURL{John} \URL{exo:friend} \LURL{Amy} . 
\\
\LURL{John} \URL{foaf:name} \literal{John}{xsd:string} . &
\LURL{Susan} \URL{exo:enrolled} \LURL{SUNYOrange} . &
\LURL{John} \URL{exo:friend} \LURL{Bill} .
\\
\LURL{John} \URL{exo:supervisor} \LURL{Len} . &
\LURL{Susan} \URL{exo:enrolled} \LURL{HudsonValley} . &
\LURL{John} \URL{exo:friend} \LURL{Len} . 
\\
\LURL{Susan} \URL{rdf:type} \URL{exo:Person} . &
\LURL{Len} \URL{exo:affiliation} \LURL{ReindeerPoly} . &
\LURL{Len} \URL{exo:friend} \LURL{Amy} . 
\\
\LURL{Susan} \URL{foaf:name} \literal{Susan}{xsd:string} . &
\LURL{Len} \URL{exo:affiliation} \LURL{SUNYOrange} . &
\LURL{Len} \URL{exo:friend} \LURL{Susan} . 
\\
\LURL{Len} \URL{rdf:type} \URL{exo:Faculty} . &
\LURL{SUNYOrange} \URL{rdf:type} \URL{exo:ResOrg} . &
\\
\LURL{Len} \URL{foaf:name} \literal{Len}{xsd:string} . &
\LURL{HudsonValley} \URL{rdf:type} \URL{exo:Uni} . &
\\

\hline
\end{tabular}
\end{figure*}

\begin{figure*}
\caption{RDFS Ontology for Example}\label{ontology}
\centering
\begin{tabular}{|l|l|}
\hline
\URL{foaf:name} \URL{rdfs:range} \URL{xsd:string} . &
\URL{exo:enrolled} \URL{rdfs:domain} \URL{exo:UniStudent} . \\
\URL{exo:UniStudent} \URL{rdfs:subClassOf} \URL{exo:Person} . &
\URL{exo:enrolled} \URL{rdfs:range} \URL{exo:Uni} . \\
\URL{exo:GrStudent} \URL{rdfs:subClassOf} \URL{exo:UniStudent} . &
\URL{exo:supervisor} \URL{rdfs:domain} \URL{exo:GrStudent} . \\
\URL{exo:Faculty} \URL{rdfs:subClassOf} \URL{exo:Person} . &
\URL{exo:supervisor} \URL{rdfs:range} \URL{exo:Faculty} . \\
\URL{exo:Uni} \URL{rdfs:subClassOf} \URL{exo:Organization} . &
\URL{exo:affiliation} \URL{rdfs:domain} \URL{exo:Person} . \\
\URL{exo:ResOrg} \URL{rdfs:subClassOf} \URL{exo:Organization} . &
\URL{exo:affiliation} \URL{rdfs:range} \URL{exo:Organization} .  \\
\hline
\end{tabular}
\end{figure*}

\begin{figure*}
\caption{Constraints and Recognition Axioms for Example}\label{constraints}
\centering
$\begin{array}{|r@{\;\;}l|r@{\;\;}l|}
\hline
1 & \URL{exo:Person} \;\AND\; \URL{exo:Organization} \Equiv \nominal{} &
7 & \URL{exo:Faculty} \Implies 
    \atmostq{5}{\inv{\URL{exo:supervisor}}}{\URL{exo:GrStudent}} \\
2 & \URL{exo:Person} \Implies  \exactly{1}{\URL{foaf:Name}} 
	\;\AND\; \all{\URL{foaf:Name}}{\URL{xsd:string}} &
8 & \URL{exo:Uni} \Implies \atleast{2}{\inv{\URL{exo:enrolled}}} \\
3 & \URL{exo:UniStudent} \Implies \atleast{1}{\URL{exo:enrolled}} 
	\;\AND\; \all{\URL{exo:enrolled}}{\URL{exo:Uni}} &
9 &\drestrict{\URL{exo:enrolled}}{\URL{exo:GrStudent}} \Implies \\[0.5ex]
4 & \URL{exo:GrStudent} \Implies \exactly{1}{\URL{exo:enrolled}} \;\AND\; 
   \all{\URL{exo:enrolled}}{\URL{exo:ResOrg}} &
  & \;\;\;\;\;\;\;\;	\URL{exo:supervisor} \comp \URL{exo:affiliation}  \\[0.5ex]
5 & \URL{exo:Faculty} \Implies \atleast{1}{\URL{exo:affiliation}} 
	\;\AND\; \all{\URL{exo:affiliation}}{\URL{exo:Uni}} &
 &\LURL{HecticStudent} \Equiv \atleast{3}{\URL{exo:enrolled}} \\
6 & \URL{exo:Faculty} \Implies \atmostq{1}{\URL{exo:affiliation}}{\URL{exo:ResOrg}} &
 &\LURL{StudentFriend} \Equiv
	\atleastq{2}{\URL{exo:friend}}{\LURL{StudentFriend}} \\
\hline
\end{array}$
\end{figure*}

Inference is the process of determining what follows from what has been
stated.  Inference ranges from simple (John is a student, students are
people, therefore John is a person) to the very complex (involving reasoning
by cases, {\em reductio ad absurdum}, or even noticing that an infinite
sequence of inferences will not produce any useful information).
Determining whether a constraint holds is just determining whether the
constraint follows from the given information.

Again, however, constraint checking is generally done with respect to
complete information.
So, to determine whether the constraint 

\begin{quote}$
\LURL{John} \in \URL{exo:Person} \;\AND\; \hfill \EXAMPLE \label{D2} \\
\hspace*{2em} \exactly{1}{\URL{foaf:name}} \;\AND\;
	\all{\URL{foaf:name}}{\URL{xsd:string}} \;\AND \\
\hspace*{2em} \exactly{1}{\URL{foaf:phone}} \;\AND\;
      \all{\URL{foaf:phone}}{\URL{xsd:string}} \;\AND \\
\hspace*{2em} \exactly{2}{\URL{exo:friend}}
$\end{quote}

\noindent is valid
in the presence of (locally) complete information such as

\begin{quote}
$\LURL{John} \in \URL{exo:Person}$ \\
\LURL{John} \URL{foaf:name} \literal{John}{xsd:string} . \\
$\LURL{John} \in \exactly{1}{\URL{foaf:name}}$ \\
\LURL{John} \URL{foaf:phone} \literal{+19085551212}{xsd:string} . \\
$\LURL{John} \in \exactly{1}{\URL{foaf:phone}}$ \\
\LURL{John} \URL{exo:friend} \LURL{Bill} . \\
\LURL{John} \URL{exo:friend} \LURL{Willy} . \\
\LURL{John} \URL{exo:friend} \LURL{Susan} . \\
$\LURL{John} \in \all{\URL{exo:friend}}{\nominal{\LURL{Bill\/},\LURL{Willy},\LURL{Susan}}}$ \\
$\notsame{\LURL{Bill\/}}{\LURL{Willy}}$ \\
$\notsame{\LURL{Bill\/}}{\LURL{Susan}}$ \\
$\notsame{\LURL{Susan}}{\LURL{Willy}}$
\end{quote}

\noindent is simply a matter of determining whether the constraint follows
from the information.  (Generally constraints like \eqref{D2} are
written to handle all the members of a class as in
\begin{quote}$
\URL{exo:Person} \Implies \\
\hspace*{2em} \exactly{1}{\URL{foaf:name}} \;\AND\;
	\all{\URL{foaf:name}}{\URL{xsd:string}} \;\AND \\
\hspace*{2em} \exactly{1}{\URL{foaf:phone}} \;\AND\;
      \all{\URL{foaf:phone}}{\URL{xsd:string}} \;\AND \\
\hspace*{2em} \exactly{2}{\URL{exo:friend}}
$\end{quote}
instead of just a single node,
but the principle is the same.)

So a way to do constraints in Description Logics is to first set up complete
information, and then just perform inference.  This approach has been
explored in the context of letting certain roles be completely specified as
in a database
\cite{patel-schneider-franconi:constraints}.   
Other approaches to
constraints in Description Logics 
\cite{debruijn-et-al:owl-flight,motik-bridging-jws,tao-et-al:constraints,donini-et-al:mknf,pascal-closed-iswc}
are considerably more complex, as they
deal with the complexities that arise when there are multiple ways to
complete the information.  However, all these approaches largely agree when
there is only a single way to complete the information.

Setting up complete information is just what was done above for closed-world
recognition, so this technique can also be used for constraint checking.
Of course, this doesn't mean that you have to implement
Description Logic inference with complete information the same way that you
need to with incomplete information.  In fact, as above, constraint
checking can be implemented as SPARQL queries.

\section{Example}

Here is a small example of how Description Logic constructs can be used for
constraint checking.  
There are three separate kinds of information in the example.
The RDF triples in Figure \ref{data} provide the data for the example.
The RDFS ontology in Figure \ref{ontology} provides the organization of the data.
The Description Logic axioms in Figure \ref{constraints} provide the
constraints to be validated against the data and the ontology and the classes
vocabulary for closed-world recognition.

The constraints are all satisfied, except for one, as follows:
\begin{enumerate}
\item Nothing can be inferred to be both a person and an organization, and
  so persons and organizations are disjoint.
\item Every object that can be inferred to be a person (i.e., students and
  faculty) has a single name provided, 
  and that name is a string, so every person has exactly one name in the
  closure. 
\item All students (and grad students) are enrolled in universities---the
  range of \URL{exo:enrolled} is \URL{exo:Uni}, which makes the typing part of
  the constraint redundant here. 
\item Reindeer Poly is not specified to be a research organization,
  so although John is enrolled exactly once the constraint on graduate
  students being enrolled in research organizations is not satisfied. 
\item All faculty (Len) are affiliated with only universities.
\item All faculty (again only Len) are affiliated with at most one research
  organization, as Reindeer Poly is not specified to be a research
  organization and Len is only affiliated with SUNY Orange and Reindeer
  Poly. 
\item All faculty supervise fewer than five graduate students, as the only
  faculty (Len) only supervises one student (John).
\item Each university (SUNY Orange, Reindeer Poly, and Hudson Valley) 
	has at least two students enrolled in it, because Amy, Bill, John,
        and Susan are different individuals.
\item For every graduate student enrollment (\URL{exo:enrolled} domain-restricted
  to \URL{exo:GrStudent}) there is a supervisor of the graduate student
  affiliated with the university.   
This constraint uses an auxiliary
  non-recursive equivalence definition.
\end{enumerate}

The only hectic student is Susan, as she is the only person with at least
three enrollments.  
However, Amy, Bill, and John all belong to \LURL{StudentFriend} because when
maximally interpreting \LURL{StudentFriend} they each have at least two
friends who belong to \LURL{StudentFriend}.  
Len does not belong to \LURL{StudentFriend} even though he has two
friends, because Susan has too few friends and cannot belong to
\LURL{StudentFriend}.   

One might want to validate that domain and range types are not
inferred, but are instead explicitly stated in the data.
This can be done by using a version of the ontology without the domain and
range statements and validating against a set of constraints that just have
the removed domain and range statements.  In the example, this would detect
that Susan was not stated to be a student, violating the domain constraint
for \URL{exo:enrolled}; 
that SUNY Orange and Reindeer Poly were not stated to
be universities, violating the range constraint for \URL{exo:enrolled};
and that Reindeer Poly was not stated to
be an organization, violating the range constraint for \URL{exo:affiliation}.
All other domain and range constraints would be satisfied, as some required
class memberships would be inferred from the subclass statements.

\section{Related Work}

The closest work in a technical sense is the work of Patel-Schneider and
Franconi \shortcite{patel-schneider-franconi:constraints}.  In that work some
properties and classes were considered as closed, which turned description
logic axioms involving those properties and classes into constraints.  RDF and
RDFS are very similar to a situation where all properties and classes
are closed.  The current paper adds the idea of closed-world recognition, which was
only implicit in the previous work, and maximal extensions, which provide a
much better treatment of recursive definitions, particularly in the monotone
case.  

The work of Motik, Horrocks, and Sattler \shortcite{motik-bridging-jws} and of Tao {\em et
  al.} \shortcite{tao-et-al:constraints} both divide up
axioms into regular axioms and constraints.  They both also permit general
Description Logic axioms, not just RDF or RDFS graphs as here.  To handle
full Description Logic information requires a much more complex
construction, involving minimal interpretations.  Neither
consider closed-world recognition.  Tao {\em et al.} use SPARQL queries as a
partial translation of their constraints and forms a basis for Stardog ICV.

The work of Sengupta, Krisnadhi, and Hitzler \shortcite{pascal-closed-iswc}
uses circumscription as the mechanism to
minimize interpretations.  It is otherwise similar to the previous efforts.
The work of Donini, Nardi, and Rosati \shortcite{donini-et-al:mknf} uses autoepistemic
constructs within axioms to model constraints, and is thus quite different
from the approach here.  
OWL Flight \cite{debruijn-et-al:owl-flight} is a subset of OWL where 
axioms are given meaning as Datalog constraints.  Again, as an expressive
Description Logic is handled the construction is more complex than the one
here.
RDFUnit \cite{rdfunit-www2014} has a component that turns RDFS
axioms and simple OWL axioms into SPARQL queries that check for
data that does not match the axiom and so is somewhat similar to this work.
However, there is no notion that RDFUnit is turning ontology axioms into
constraints that cover the entire meaning of the axiom.  

Shex \cite{shex-definition} uses very different mechanisms.  It builds up
shapes, which are akin to definitions of classes, and gives them meaning by
a translation into a recursive extension of Z over an abstraction of RDF
graphs.  Entire documents or document portions are then matched against
these shapes. 

\section{The Details, but Not All the Details}

\subsection{Description Logic Semantics}

The semantics of Description Logics are generally given as a model theory,
as for OWL DL \cite{motik-et-al:owl-ss-fs}.  OWL DL has a complex
semantics, as far as Description Logics go, to cover all its constructs
and to make it more compatible with RDF.  The semantics here will 
follow the semantics of OWL, with the exception that any property can have
both individuals, e.g., \LURL{John}, and data values, e.g.,
\literal{John}{xsd:string}, as values.  

The fundamental building block of Description Logics
semantics is interpretations, which provide a meaning for the primitive
constructs in terms of a particular domain of discourse.
The meaning of an individual name, such as \LURL{John}, is an element
of this domain, here that element that we think of as being John.
Literal values, such as \literal{John}{xsd:string}, are
treated specially---their meaning is determined by their datatype.
The meaning (or interpretation) of a named concept such as \URL{exo:person}, is a
subset of the  domain, here those individuals that we might think of as people.
The meaning of a named property, such as \URL{exo:friend}, is a set of pairs
over the domain, here those pairs that we might think of as being the
friend-of relationship.  The meaning of non-primitive constructs, such as
the description $\exactly{2}{\URL{exo:friend}}$, are built up from these
primitives, resulting here in the set of domain elements that are related to
exactly two domain elements via the meaning of \URL{exo:friend}.  

Axioms, such as $\LURL{John} \in \URL{exo:Person}$, are true precisely when
the meaning of their parts 
satisfies a particular relationship, here that the meaning of \LURL{John} is
an element of the meaning of \URL{exo:Person}.
There are some other aspects to this simple story, to handle the differences
between individuals, e.g., \LURL{John}, and data values, e.g.,
\literal{John}{xsd:string}, and to make reasoning over
some constructs easier.
A Description Logic model of a set of axioms (including what we might call
facts), is then just an interpretation that makes all the axioms true.

\subsection{Canonical Interpretations of RDF Graphs}

In an interpretation everything is specified, so each interpretation has
complete information.  The basic idea is thus to construct an interpretation
making just the triples in an RDF graph true and then work with that single
interpretation.  In this way information that is absent from the RDF graph
is considered false.

We can think of most RDF graphs as sets of Description Logic axioms,
particularly as we are ignoring the common Description Logic division of
properties into properties that have objects that are individuals and
properties that have objects that are data values.\footnote{This division
  has been made so that reasoners do not have to worry about data values
  having properties, but if we are constructing models this is not a
  problem.  It is easy to revise the treatment here to bring back this
  division.}  This correspondence breaks down in two areas: 1/ when the
built-in RDF and RDFS vocabulary is used in unusual ways (e.g., making
\URL{rdfs:subClassOf} a sub-property of \URL{rdfs:subPropertyOf}), and 2/ if
reasoning about individuals can affect reasoning about classes (e.g.,
forcing two individuals that are also classes to be the same).  The abuse
and extension of the built-in RDF and RDFS vocabulary is rare, so we just
exclude these RDF graphs from our account.  (Particular extensions of
  the RDF and RDFS vocabulary could 
  be built in to an extension of the approach given here.)  In RDF and RDFS,
but not in OWL, reasoning about individuals can affect reasoning about
classes.  This is also rare, so we
will not handle these inferences.

\begin{definition}
Given an RDF graph $G$ with no ill-formed literals and no triples stating
membership in a datatype, we construct the
canonical Description Logic interpretation of $G$ as follows.
\begin{enumerate}
\item Datatypes are formed for all the datatypes in the graph, and
  given meaning in the usual way.  
\item The domain of the interpretation consists of the non-literal nodes of
  the RDF graph plus the properties in the graph and the mapping for nodes
  is the identity mapping.  (One might think that this
  is not an appropriate way to construct an interpretation, as it sets the
  meaning of \LURL{John} to \LURL{John}, not anything that we might think of
  as being John, but as far as the formal machinery is concerned, the actual
  domain elements are not important.)
\item The set of literal values is constructed in the usual way from the
  datatypes. An extra copy of the integers is added to ensure an infinite
  number of literal values.
\item Classes are formed for each node in the graph that has an
  \URL{rdf:type} link with it as an object or belongs to \URL{rdfs:Class}
  and their extensions are the
  set of nodes for which \URL{rdf:type} triples link them to the class.  
\item Properties (note that we are ignoring the
  Description Logic division of properties) are formed for each
  predicate in the graph and also for each node that belongs to \URL{rdf:Property},
  and their extensions are the set of pairs taken from triples in the graph with the property as predicate.
\end{enumerate}
\end{definition}

So from the initial RDF graph in this paper, we end up with a canonical
interpretation with six domain elements, \LURL{John}, 
\LURL{Bill\/}, \LURL{Willy}, \URL{foaf:name}, \URL{foaf:phone}, and
\URL{exo:friend}.  
The interpretation of \URL{foaf:name} consists of just 
$\left< \LURL{John} ,  \verb|"|\URL{John}\verb|"| \right>$
The interpretation of \URL{foaf:phone} consists of just 
$\left< \LURL{John} ,  \verb|"|\URL{+19085551212}\verb|"| \right>$.
The interpretation of \URL{exo:friend} consists of 
$\left< \LURL{John} , \LURL{Bill\/} \right>$ and
$\left< \LURL{John} , \LURL{Willy} \right>$.

For constraints and descriptions that use only vocabulary in the RDF graph
all we do is work with this
interpretation and consider whether the constraint axiom is true in this
interpretation so the development is easy.  It is obvious that \LURL{John}
belongs to the interpretation of the first Description Logic description
given above, as expected.  

Evaluating constraints on the canonical interpretation of a graph is
essentially the same as evaluating them on the graph itself.  Systems that
evaluate constraints on an RDF graph, like ShEx, thus work in a manner
very similar to the approach taken here.

\subsection{Extending to New Classes}

For closed-world recognition, it is useful to define new classes, as in 
\begin{quote}
$\LURL{PurePerson} \Equiv \atleast{1}{\URL{exo:friend}} \;\AND\; \\
\hspace*{3em}\all{\URL{exo:friend}}{\LURL{PurePerson}}$
\end{quote}
There are several possibilities for the meaning of new
classes that are recursively defined.
The new classes could be interpreted as broadly as 
possible, as narrowly as possible, or in any consistent manner.

It appears in ShEx that such classes as
to be interpreted as broadly as possible.
For example, in the RDF graph
\begin{quote}
\LURL{John} \URL{exo:friend} \LURL{Bill\/} . \\
\LURL{Bill\/} \URL{exo:friend} \LURL{John} .
\end{quote}
\noindent the ShEx approach would be that 
both \LURL{John} and \LURL{Bill} belong to \LURL{PurePerson}.
We will take this approach here and interpret new
classes as broadly as possible

New classes are handled by considering extensions of the interpretations
defined as above.  An extension of an interpretation is a new interpretation
1/ with the same domain as the original interpretation, and 
2/ that has the same meaning for all individuals, named classes, and
  named properties in the original interpretation.
The extension is
allowed to have new named classes, but not new named properties or new
individuals.  New individuals are not allowed because some Description
Logic constructs are sensitive to the set of individuals.  New properties
are not allowed because they may increase the computational complexity of
closed-world recognition.

An interpretation is an extended canonical model of an
RDF graph with respect to a set of constraints if it is an extension of the
canonical model of the RDF graph and is a model of the constraints.

To interpret recursively defined classes as broadly as possible not all
extended canonical models are considered, only maximal ones.  An
model is maximal among a set of models if 
there is no other model in the set that 
1/ interprets all classes as supersets of their
  interpretation in the maximal model, and
2/ interprets at least one class as a strict superset of its
  interpretation in the maximal model.

An individual is recognized as belonging to a description if its
interpretation belongs to the interpretation of the description in all
maximal extended canonical models.  

It turns out that there is only one (up to isomorphism) maximal extended
canonical model of the above definition of \LURL{PurePerson}.  In this model both
\LURL{John} and \LURL{Bill\/} are in the extension of \LURL{PurePerson}.  If all new classes are monotone in all the other new
classes (i.e., if the extension of some class grows then no
other class extensions shrink) then there is always exactly one maximal
extension.\footnote{Proof sketches of claims here and later in the paper are in  
the appendix of this extended version.}

\subsection{RDF and RDFS Semantics}

So everything looks fine.  We go from an RDF graph to a slightly modified
Description Logic interpretation and from there perform constraint
checking by determining whether a set of Description Logic axioms are
satisfied in the interpretation or in a set of maximal extensions of the
interpretation. 
We can also perform closed-world recognition by determining the
interpretation of the new defined classes in the axioms and these classes can
even be defined recursively.

However, there is one missing part of the story.  If the RDF graph includes
triples that trigger RDF or RDF inferences that are not already in the RDF
graph the interpretation will not look like an RDF (or RDFS)
interpretation.  For example, if the RDF graph is 
\begin{quote}
\LURL{John} \URL{rdf:type} \URL{exo:Student} . \\
\URL{exo:Student} \URL{rdfs:subClassOf} \URL{exo:Person} . 
\end{quote}
our canonical interpretation states that \LURL{John} does not belong to
\URL{exo:Person}, which goes against the RDFS meaning of the above graph.

Fortunately, it is relatively easy to recover from this problem.  All that
is needed is to add all the RDF (or RDFS) consequences to the graph.  Yes,
there are an infinite number of these consequences, but our formal
development does not care whether the graph is finite or infinite.  For
complexity analysis and implementation it is not hard to come up with a
finite representation of these consequences, like the one initially done by
ter Horst \shortcite{ter-horst-rdfs}, and make the minor fixups needed to
determine correct answers from the answers gleaned from this finite
approximation.

This all works because the RDF (or RDFS) consequences of an RDF graph can be 
represented as an RDF graph.  OWL consequences cannot be so
represented, as the consequences in OWL can be disjunctive.  This
requires working with minimal equality and minimal models or some other way
to single out only the desired interpretations as in previous work on
Description Logic constraints, making the formal 
development much harder and presenting many more choices that have to be
justified. 

\section{Complexity}

It is easy to see that checking axioms against an interpretation is
polynomial, as long as there is no new vocabulary in the axioms or no
recursive definitions.
The formulae corresponding to the axioms are just model checked against the
interpretation. 

If there are monotone recursive definitions then checking constraints and
performing closed-world recognition can be done using techniques from 
Datalog, such as magic sets.
For example, the extension of a recursively-defined class can first be
computed ignoring the recursive portion.  Violations of the recursive
portion can then be checked and objects iteratively removed from the class. 

These techniques cannot be used for non-monotone recursive definitions,
as expanding one class or property might reduce another.

\section{Implementation}

The work of Tao {\em et al.} \shortcite{tao-et-al:constraints} shows that the
standard Description Logic constraints can be partly implemented as SPARQL
queries when no new vocabulary is used.
Tao {\em et al.}~worked in a general OWL setting, where their approach is sound
but not complete, but in an RDF setting the approach is 
both sound and complete, because there is only a single model that needs to
be considered. 
This approach forms the basis of Stardog ICV \cite{stardog-icv}.
Indeed Stardog ICV is an implementation of the approach described in this
paper showing how the approach to constraints here can be implemented by a
translation to SPARQL.
The work here can thus also be thought of as a simpler definition of the
underpinning of Stardog ICV.
Recent work at Mannheim by Thomas Bosch
(see \URL{https://github.com/boschthomas/OWL2-SPIN-Mapping}) 
translates OWL
descriptions interpreted as constraints into SPARQL using a similar
approach, providing a different implementation.

Non-recursive closed-world recognition can be handled by using nested or
repeated SPARQL queries.  Monotone recursive closed-world recognition can be
implemented using Datalog techniques.  Non-monotone recursive
closed-world recognition is more complex and cannot be handled in the same
way.  This indicates that excluding non-monotone recursive closed-world
recognition could be a reasonable stance to take.

\section{Conclusion}

Description Logics can indeed be used for both the syntax and semantics
of constraint checking and closed-world recognition in RDF, by employing
an analogue of model checking, and much of both constraint checking and
closed-world recognition can be effectively implemented using a translation
to SPARQL queries. 
The main difference between closed-world recognition and constraint checking is
that the former either has no axioms or only uses axioms defining names that
do not occur in the RDF graph whereas constraint checking uses axioms that
relate concepts appearing in the RDF graph to descriptions.

By restricting our information to be RDF or RDFS, i.e, working in situations
where there is a unique minimal model, we obtain a simpler
formulation, easy implementation, and good performance as compared to
previous work in this area.
The approach here can be easily extended to other subsets of OWL that have
a unique minimal model.  An OWL profile with this property is OWL RL
\cite{owl-2-profiles}.   

\bibliographystyle{named}
\bibliography{publications}

\appendix

\section{Appendix}

\bigskip

\subsection{Description Logic Semantics}

\medskip

This is a condensed version of Description Logic semantics, largely taken
from the OWL 2 semantics \cite{motik-et-al:owl-ss-fs}.  
Some parts of the
semantics have been removed (including facets
and naming) so that this account is easier to read.   As well, the division
of properties between object properties and data properties has been
removed so that RDF properties that have both objects and data values as
objects can be handled.  Many notions from RDF and Description
Logics will be used without definition.

\begin{definition}
A {\em datatype} consists of 
the name of the datatype, which is an IRI;
the set of values for the datatype,
and a partial mapping from strings to these values.
\end{definition}

\begin{definition}
A  (class, property, and individual) {\em vocabulary} is a tuple
$\left< V_C, V_P, V_I \right>$
where $V_C$, the classes, 
$V_P$, the properties, 
and $V_I$, the individuals, 
are each sets of IRIs and blank nodes,
\end{definition}

\noindent Note that there is no requirement that the classes, properties, and
individuals in a vocabulary be pairwise disjoint.

\begin{definition}
Given a vocabulary $V$ and set of datatypes $D$, 
an {\em interpretation} is a tuple
$\left< \Delta_I, \Delta_D, \cdot^{C}, \cdot^{P}, \cdot^{I} \right>$,
where
\begin{itemize}
\item $\Delta_I$ is a non-empty set of objects, the domain of the interpretation,
\item $\Delta_D$ is a set of data values, containing at least all the values
	for datatypes in $D$,
\item $\cdot^{C}$ maps $V_C$ into subsets of $\Delta_I$,
\item $\cdot^{P}$ maps $V_P$ into subsets of 
	$\Delta_I \times ( \Delta_I \cup \Delta_D )$, and
\item $\cdot^{I}$ maps $V_I$ into elements of $\Delta_I$.
\end{itemize}
\end{definition}

\noindent The semantics also uses $\cdot^{DT}$, which maps dataypes in $D$
into the set of their values as specified by the datatype, and 
$\cdot^{LT}$, which maps literals into their values as specified by the datatypes.
$\cdot^{C}$ and $\cdot^{P}$ are extended to class expressions
and property expressions in the usual way.

\begin{definition}
A Description Logic axiom (i.e., an OWL axiom) is {\em true} in an interpretation
in the usual way, with appropriate modifications made to eliminate the
distinction between object and data properties.
A {\em model} of a set of Description Logic axioms is an interpretation that makes
all the axioms true. 
\end{definition}

\bigskip

\subsection{Same-Vocabulary Constraints}

\medskip

\begin{definition}
The {\em vocabulary} for an RDF graph $G$ is 
$V = \left< V_C, V_P, V_I \right>$
where 
\begin{align*}
V_C = & \{ C \;|\; \exists s \; \left< s, \URL{rdf:type} , C \right> \in G \} \;\cup \\
      &	\{ C \;|\; \left< C , \URL{rdf:type} , \URL{rdfs:Class} \right> \in G \}, \\
V_P = & \{ P \;|\; \exists x,o \; \left< s , P , o \right> \in G \} \;\cup \\
      & \{ P \;|\; \left< P , \URL{rdf:type} , \URL{rdf:Property} \right> \in G \},
\end{align*}
and $V_I$ is the set of non-literal nodes in $G$.
\end{definition}

\begin{definition}
Given a set of datatypes $D$,
the {\em canonical interpretation}
for an RDF graph $G$ using only datatypes in $D$ is 
the interpretation 
$I = \left< \Delta_I, \Delta_D, \cdot^{C}, \cdot^{P}, \cdot^{I} \right>$
over the vocabulary of $G$, $\left< V_C, V_P, V_I \right>$, and the datatypes in $D$,
where
\begin{itemize}
\item $\Delta_I = V_C \cup V_P \cup V_I$,
\item $\Delta_D$ is the union of all the values for datatypes in $D$
	disjointly unioned with a copy of the integers,
\item $c^{C} = \{ s | \exists \left< s , \URL{rdf:type} , c \right> \in G \}$, for $c \in V_C$,
\item $p^{P} = \{ \left< s,o \right> | \exists \left< s , p , o \right> \in G \}$, for $p \in V_P$, and
\item $i^{I} = i$, for $i \in V_I$.
\end{itemize}
\end{definition}

The canonical interpretation of $G$ includes all classes and properties in
$G$ as individuals.  It also constructs Description Logic classes and
properties for the built-in RDF and RDFS classes and properties such as
\URL{rdfs:Class}, \URL{rdf:type}, and \URL{rdfs:subClassOf}.
This is not exactly what might be expected, 
does create a reasonable interpretation that matches RDF and RDFS intuitions
closely.

\begin{theorem}
Given a set of datatypes $D$ and
an RDF graph $G$ using only these datatypes,
if $G$ is closed under the RDF (RDFS) rules of inference
then a minor adjustment to the canonical interpretation of $G$ is a model of
$G$ under the RDF (RDFS) semantics.
\proof
Adjustments first have to be made to the canonical model to turn it into an
actual RDF interpretation.  The RDF domain is the union of the objects and
data values of the canonical interpretation.  Literals are placed into the
datatypes they belong to.  

The proof is then via a simple case-by-case analysis of each semantic condition
on RDF (RDFS) models.
\end{theorem}

\begin{definition}
Given a set of datatypes $D$,
an RDF graph $G$ with vocabulary $V$,
and a set of Description Logic axioms $C$ over $V$ acting as constraints, 
$C$ is {\em satisfied} by $G$ iff
the canonical interpretation of $G$ is a model for $C$.
\end{definition}

Note that in many cases where $C$ is satisfied by $G$,
$C$ will not follow from $G$.  For example 
\begin{quote}
$\left\{ \URL{ex:Person} \Implies \atmost{1}{\URL{foaf:name}} \right\}$
\end{quote}
is satisfied by but does not follow from
\begin{quote}
\LURL{John} \URL{rdf:type} \URL{ex:Person} . \\
\LURL{John} \URL{foaf:name} \literal{John}{xsd:string} . 
\end{quote}

\begin{theorem}
Given an RDF graph $G$ with vocabulary $V$ 
and a set of constraints $C$ over $V$ and with no blank nodes,
checking whether $G$ satisfies $C$ can be done in polynomial time for
most Description Logics, including OWL.
\proofsketch
For most axioms checking can be reduced to checking inclusion relationships
between descriptions in the canonical model.  Determining the extension of a
description in a model involves checking some local conditions, which can be
easily done in polynomial time because there are no choices to be made in
the model, which specifies the extension of all named classes and properties.
Some axioms are not reducible to checking inclusion relationships (e.g, key
axioms) but are similarly easy to check in a model.  The prohibition of
blank nodes is to prevent the creation of sets of individual axioms that
require checking for the presence of graph structures in the canonical model.
\end{theorem}

Blank nodes do not cause a problem here because the only blank nodes allowed
in the constraints are blank nodes that also are in the graph, and these
blank nodes are treated just the same as if they are IRIs.  The price paid
is that it is not possible to use blank nodes in the constraints to perform
structure matching, for example to see if there is some individual that is
related to another particular individual via two separate role chains.

\begin{theorem}
Given an RDF graph $G$ and vocabulary $V$
and a set of constraints $C$ over $V$
determining whether $G$ satisfies $C$ 
can be done by a number of SPARQL queries on $G$  polynomial in the size of $C$.
\proofsketch
Description Logic description or property expressions have obvious
translations to SPARQL queries that when run on $G$ produce the extension of
the description.   The translations are the ones used in Stardog ICV
as described by Tao {\em et al.} \shortcite{tao-et-al:constraints}.
The Description Logic axioms that check the relationship between two
description or property expressions can be checked by creating a SPARQL
query that is empty if and only if the axioms is satisfied by $G$, 
again as done by Tao {\em et al.}
Other Description Logic axioms, such as transitivity and keys can be treated
in similar ways.
\end{theorem}

\bigskip

\subsection{Class-Extended Constraints}

\medskip

\begin{definition}
Given a vocabulary $V = \left< V_C, V_P, V_I \right>$ and a set of datatypes $D$,
$I' = \left< \Delta'_I, \Delta'_D, \cdot^{C'}, \cdot^{P'}, \cdot^{I'} \right>$,
an interpretation over vocabulary $V' = \left< V'_C, V'_P, V'_I \right>$,
is an {\em extension} of an interpretation 
$I = \left< \Delta_I, \Delta_D, \cdot^{C}, \cdot^{P}, \cdot^{I} \right>$
over $V$ iff
\begin{itemize}
\item $V_C \subseteq V'_C$, $V_P = V'_P$, $V_I = V'_I$,
\item $\Delta_I = \Delta'_I$, $\Delta_D = \Delta'_D$,
\item $c^{C} = c^{C'}$ for $c \in V_C$,
\item $p^{P} = p^{P'}$ for $p \in V_P$, and
\item $i^{I} = i^{I'}$ for $i \in V_I$.
\end{itemize}
\end{definition}

\noindent Note that only the class vocabulary can be extended; the
property and individual vocabularies remain the same.  Also note that the
domain is unchanged.

\begin{theorem}
If $I$ is a model of the RDF graph $G$ with vocabulary $V$ and $I'$ is an extension of $I$, then
$I'$ is a model of $G$.
\proof
$I$ and $I'$ only differ on the class vocabulary outside $V$,
which does not affect whether an interpretation is a model of $G$ even for
Description Logic constructs that are sensitive to the individual vocabulary.
\end{theorem}

\begin{definition}
Given a set of datatypes $D$,
an RDF graph $G$ with vocabulary $V$,
and a set of Description Logic axioms $C$
whose properties (individuals) are all properties (individuals) from $V$,
$C$ is {\em satisfied} by $G$ iff
there is an extension of the canonical interpretation of $G$ that is a model
for $C$. 
\end{definition}

\noindent Note that each class in $C$ that
has an equality definition that does not directly or indirectly refer to itself
has the same class extension in each extension of the canonical
interpretation of $G$ that is a model for $C$.

\begin{definition}
Given a set of datatypes $D$, 
an RDF graph $G$ with vocabulary $V$,
and a set of Description Logic axioms $C$ 
whose properties (individuals) are all properties (individuals) from $V$,
$o$ is in the {\em closed-world class extension} of $c$ for $o$ an
individual in $G$ and $c$ a class in $G$ or $C$ iff
$m^{I}(o) \in m^{C}(c)$ in each extension, $m$, of the canonical
interpretation of $G$ that is a model for $C$. 
\end{definition}

\medskip

\begin{definition}
Given a vocabulary $V = \left< V_C, V_P, V_I \right>$ and a set of datatypes $D$,
$I' = \left< \Delta'_I, \Delta'_D, \cdot^{C'}, \cdot^{P'}, \cdot^{I'} \right>$,
an interpretation over $V$
is bigger than or equal to an interpretation
$I = \left< \Delta_I, \Delta_D, \cdot^{C}, \cdot^{P}, \cdot^{I} \right>$
over $V$ iff
$c^{C} \subseteq c^{C'}$ for all $c \in V_C$,
$p^{P} = p^{P'}$ for all $p \in V_P$, and
$i^{I} = i^{I'}$ for all $i \in V_I$.
\end{definition}

\begin{definition}
Given a set of datatypes $D$,
an RDF graph $G$ with vocabulary $V$,
and a set of Description Logic axioms $C$
whose properties (individuals) are all properties (individuals) from $V$,
an interpretation 
$I = \left< \Delta_I, \Delta_D, \cdot^{C}, \cdot^{P}, \cdot^{I} \right>$ 
over vocabulary $V'$ 
is a {\em maximal extension model} of $G$ and $C$
iff
\begin{itemize}
\item $I$ is an extension of the canonical interpretation of $G$,
\item $I$ is a model of $C$, and
\item there is no model of $G$ and $C$ that is bigger than $I$.
\end{itemize}
\end{definition}

\begin{definition}
Given a set of datatypes $D$, 
an RDF graph $G$ with vocabulary $V$,
and a set of Description Logic axioms $C$
whose properties (individuals) are all properties (individuals) from $V$,
$o$ is in the {\em maximal closed-world class extension} of $c$ for $o$ and $c$
nodes in $G$ or $C$ iff
$m^{I}(o) \in m^{C}(c)$ in each maximal extension model, $m$, of $G$ and $C$.
\end{definition}

\begin{definition}
Given a set of datatypes $D$, 
an RDF graph $G$ with vocabulary $V$, 
a set of constraints $C$
whose properties (individuals) are all properties (individuals) from $V$,
and two class names in $C$ but not in $V$, $C_1$ and $C_2$,
$C_1$ is {\em monotone} with respect to $C_2$
if whenever $I_1$ and $I_2$ are models of $G$ that satisfy $C$ and have the
same extension for all classes in $C$ except for $C_1$ and
$C_2$ then if $I_1^C(C_2) \subseteq I_2^C(C_2)$ then 
$I_1^C(C_1) \subseteq I_2^C(C_1)$.
\end{definition}

\begin{definition}
Given a set of datatypes $D$, 
an RDF graph $G$ with vocabulary $V$, and
a set of constraints $C$
whose properties (individuals) are all properties (individuals) from $V$,
then $C$ is {\em monotone} iff all
class names in $C$ but not in $V$
are monotone with respect to all
class names in $C$ but not in $V$.
\end{definition}

\medskip

\begin{theorem}
Given a set of datatypes $D$, 
an RDF graph $G$ with vocabulary $V$, and
a set of monotone constraints $C$ over $G$ and $V$, 
there is one maximal extension for $G$ and $C$ up to isomorphism.
\proof
Take a model of $G$ and $C$ that has the biggest extension for some class in $C$ but not in $V$.
This model must also have maximal extensions for all the other classes in $C$ but not in $V$ because otherwise $C$ would not be monotone.
This model is thus, up to isomorphism, maximal.
\end{theorem}

\end{document}